\documentclass[a4paper,twoside]{article}

\usepackage{epsfig}
\usepackage{subcaption}
\usepackage{calc}
\usepackage{amssymb}
\usepackage{amstext}
\usepackage{amsmath}
\usepackage{amsthm}
\usepackage{multicol}
\usepackage{pslatex}
\usepackage{apalike}
\usepackage{algorithm2e}
\usepackage{graphicx}
\usepackage[bottom]{footmisc}
\usepackage{adjustbox}
\usepackage{hyperref}

\usepackage{SCITEPRESS}     

\begin{document}

\title{Multi-view Inversion for 3D-aware Generative Adversarial Networks}

\author{
\authorname{Florian Barthel\sup{1}\sup{2}\orcidAuthor{0009-0004-7264-1672}, 
Anna Hilsmann\sup{2}\orcidAuthor{0000-0002-2086-0951} 
and Peter Eisert\sup{1}\sup{2}\orcidAuthor{0000-0001-8378-4805}}
\affiliation{\sup{1}Humboldt Universit\"at zu Berlin, Berlin, Germany}
\affiliation{\sup{2}Fraunhofer HHI, Berlin, Germany}
\email{florian.tim.barthel@hhi.fraunhofer.de}
}



\abstract{Current 3D GAN inversion methods for human heads typically use only one single frontal image to reconstruct the whole 3D head model. This leaves out meaningful information when multi-view data or dynamic videos are available. Our method builds on existing state-of-the-art 3D GAN inversion techniques to allow for consistent and simultaneous inversion of multiple views of the same subject. We employ a multi-latent extension to handle inconsistencies present in dynamic face videos to re-synthesize consistent 3D representations from the sequence. As our method uses additional information about the target subject, we observe significant enhancements in both geometric accuracy and image quality, particularly when rendering from wide viewing angles. Moreover, we demonstrate the editability of our inverted 3D renderings, which distinguishes them from NeRF-based scene reconstructions.
}

\onecolumn \maketitle \normalsize \setcounter{footnote}{0} \vfill

\section{\uppercase{Introduction}}
\label{sec:introduction}

With recent advancements in 3D-aware image synthesis using NeRF-based Generative Adversarial Networks (GANs) \cite{gu2021stylenerf,chan_pi-gan_2021,chan_efficient_2022}, multiple methods have been proposed to re-synthesize 3D renderings for given target images using GANs \cite{xie_high-fidelity_2023,xia_gan_2022,pixel2style2pixel,an_panohead_2023}. This process called 3D GAN inversion typically uses a single input image to reconstruct / invert a 3D representation. 
3D GAN inversion methods are especially popular as they facilitate object editing and manipulation of object attributes, such as shape, texture, or pose.


In this process, a target image is first inverted, by optimizing a location in the latent space of the GAN that re-synthesizes the input image. Subsequently, the output image can be manipulated using methods like GAN-Control \cite{shoshan_gan-control_2021} or GANSpace \cite{harkonen_ganspace_2020}. This opens up a wide range of possible applications for various industries.

When inverting human face images, the input target image usually shows the head from a frontal view. This specific viewpoint provides the inversion method with a substantial amount of information about the person's identity and important cues related to the coarse geometry. However, it misses a significant amount of details that are only visible from other viewing directions.
This leads to inaccurate representations for large viewing angles, as the GAN has to extrapolate information from training data to cover regions that are not visible in the target image. In addition, to reconstruct the geometry of a given face from a frontal viewpoint, the inversion method has to rely on the silhouette, shading, and shadows, which might be ambiguous under certain conditions.

To overcome the mentioned limitations, we present a novel 3D GAN inversion method that allows to re-synthesize consistent 3D representations from multiple views at once. Only using a short video clip, showing a head turning from one side to the other, we are able to reconstruct an accurate 3D re-synthesis from arbitrary viewing angles. We show that our method significantly improves the geometry and especially the image quality when rendering from large viewing angles, compared to prior 3D GAN inversion methods. We compare our method to state-of-the-art 3D GAN inversion methods, outperforming them in terms of reconstruction quality, and demonstrate editing capabilities of the consistent representation.

Our project page can be found here: \url{https://florian-barthel.github.io/multiview-inversion}.

\section{\uppercase{Related Work}}
\label{sec:related_work}

\subsection{3D-aware GANs}
Following the success of the StyleGAN model for 2D image synthesis \cite{karras_style-based_2019,karras_analyzing_2020,karras_alias-free_2021}, numerous architectures have been developed to enable 3D-aware image synthesis using StyleGAN as the backbone \cite{niemeyer_giraffe_2021,shi_lifting_2021,chan_pi-gan_2021,brehm_controlling_2022}. The Efficient Geometry-aware 3D GAN (EG3D) \cite{chan_efficient_2022} stands out in particular due to its high resolution and good image quality. 
EG3D uses a StyleGAN2 generator to synthesize 2D features, which are then orthogonally aligned in a 3D space. From this structure, called Tri-Plane, rays can be computed using a NeRF renderer \cite{barron_mip-nerf_2021} to create novel 3D views with consistent geometry.

\subsection{2D GAN Inversion}
GAN inversion seeks to identify the precise latent location in the latent space or generator configuration that best re-synthesizes a particular target image.
Generally, GAN inversion methods can be separated into three groups: optimization-based, encoder-based, and hybrid approaches. 
\\

\noindent {\bf Optimization:}
For optimization-based methods, a latent vector $z \in \mathcal{Z}$ is optimized with the goal of synthesizing an image that closely resembles the target image. This can be done via gradient descent using differentiable loss functions. These loss functions can range from simple pixel-wise $L_1$ or $L_2$ losses to more complex perceptual losses, such as the $LPIPS$ distance \cite{lpips} or an $ID$ loss \cite{arcface}. Starting from a random latent vector $z$, the position is optimized as follows:

\begin{equation}
\label{eq1}
    z_{n+1} = z_n - \alpha \frac{d}{dz_n} \mathcal{L}(G(z_n), t).
\end{equation}

Here, $\alpha$, $G$ and $t$ denote the learning rate, the generator and the target image, respectively. $\mathcal{L}$ describes the loss function that measures the image similarity.

Specifically for StyleGAN inversion, it is common to optimize a latent vector within the $\mathcal{W}$ space. This latent space is created by StyleGAN's mapping network. Compared to the input space $\mathcal{Z}$ it is less entangled, which means that similar output images have close spatial locations in $\mathcal{W}$. This helps the inversion process, as the update steps have a more guided direction towards the target. 

Recent studies \cite{abdal_image2stylegan_2019,abdal_image2stylegan_2020} also show that optimization in the $\mathcal{W+}$ space can improve the inversion quality even further. $\mathcal{W+}$ describes the $\mathcal{W}$ space with additional block-wise modifications, i.e., $\mathcal{W+}$ inversion directly optimizes the input for each StyleGAN block separately. This provides considerably more parameters,  facilitating an accurate inversion.
\\

\noindent {\bf Encoder:}
Encoder-based inversion methods such as \cite{enc_1,enc_2,enc_3,pixel2style2pixel} trade pre-training time for inference time by previously training an encoder on the latent space. This results in an image-to-image architecture, where a target image is first mapped to the latent space and afterwards re-synthesized by the generator. This enables real time GAN inversion, whereas optimization-based methods can take up several minutes for one target image. While optimization-based methods generally show better results than encoder-methods, some recent methods \cite{bhattarai_triplanenet_2023} achieve equivalent results by also predicting an offset for the generator weights.
\\

\noindent {\bf Hybrid:}
Finally, multiple methods \cite{hybrid_3,hybrid_1,hybrid_2} have combined both optimization- and encoder-based inversion methods. In such a case, a target image is first encoded into the latent space and then the resulting latent vector is further optimized via gradient descent. Depending on the use case, hybrid methods can yield a good trade-off between inversion quality and computing time.

\subsection{Pivotal Tuning Inversion}
After finding the location of a latent vector that re-synthesizes the target image, it can be beneficial to fine-tune the generator weights $\theta$ with respect to the target image. This method called Pivotal Tuning Inversion (PTI) \cite{roich_pivotal_2021} simultaneously improves the similarity to the target, while also improving the editability. Similarly to the equation (\ref{eq1}), PTI can be expressed as follows:

\begin{equation}
\label{eq2}
    \theta_{n+1} = \theta_n - \alpha \frac{d}{d\theta_n} \mathcal{L}(G(z;\theta_n), t).
\end{equation}

\subsection{3D GAN Inversion}
Given a 3D-aware GAN model, we can apply all the aforementioned 2D GAN inversion methods to perform 3D GAN inversion. For instance, PanoHead \cite{an_panohead_2023} uses $\mathcal{W+}$ inversion followed by PTI to successfully reconstruct a target image. Another recent method is TriplaneNet \cite{bhattarai_triplanenet_2023}, which learns an encoder network to predict both a latent vector and offsets for the generator weights to produce state-of-the-art 3D GAN inversions. Other methods, such as \cite{symetry_prior} or \cite{xie_high-fidelity_2023}, first generate multiple pseudo views from one single target and then perform optimization on the real and the synthesized images simultaneously.

\subsection{Video to 3D Representation}
With the growing popularity of 3D scene reconstruction using NeRFs, multiple works \cite{park_nerfies_2021,park_hypernerf_2021,xu_4k4d_2023} have attempted 3D reconstruction from dynamic videos. One of the most popular methods is HyperNeRF \cite{park_hypernerf_2021}. It learns a high dimensional deformation field for the rays inside NeRF's renderer. This enables a consistent 3D reconstruction from dynamic data as all inconsistencies can be compensated by the deformation field.

TriplaneNet \cite{bhattarai_triplanenet_2023} proposes another video to 3D method, where each frame is inverted separately using an encoder-based 3D GAN inversion method. However, instead of returning one consistent 3D model, TriplaneNet's approach returns a separate model for each frame.

Our approach differs from both aforementioned methods. Instead of applying a NeRF-based scene reconstruction algorithm, we invert a 3D-aware GAN, i.e.~the EG3D \cite{chan_efficient_2022}. This enables us to edit the re-synthesized image afterwards. Contrary to TriplaneNet's video to 3D approach, our method returns one coherent 3D model that yields good inversion quality from all viewing angles, instead of synthesizing one representation for each frame. 

Our main contributions in this paper are:

\begin{itemize}
    \item a \textit{multi-view} inversion optimization algorithm for synthesizing consistent 3D renderings using multiple viewpoints simultaneously,
    \item a \textit{multi-latent} inversion optimization algorithm to improve 3D inversion capabilities when utilizing dynamic videos,
    \item three consistency regularization methods for improving the multi-latent optimization. 
\end{itemize}

\section{\uppercase{Method}}
\label{sec:method}

For our consistent multi-view GAN inversion method, we chose to use an optimization-based approach. This decision has two reasons. (i) We focus on inversion quality, rather than computing time; and (ii) it makes our method applicable to any 3D-aware GAN, regardless of the training data, and without any prior training of an encoder network.

Our method consists of two parts. First, we propose a multi-view inversion method, where we build on existing GAN inversion methods to be applicable to multiple target images, creating consistent representations across views (section \ref{multi-view-single-w}). Afterwards, we fine-tune multiple latent vectors, each responsible for a certain viewing angle, to account for small dynamic inconsistencies in a dynamic input video sequence (section \ref{multi_latent_opt}). 

\subsection{Multi-view Optimization}
\label{multi-view-single-w}
Given a target video of a human face, we sample $N$ target frames equally distributed from the leftmost to the rightmost camera perspective. To do so, we estimate all camera parameters using the algorithm from \cite{deng2019accurate}, and select frames that are closest to precomputed directions.

\begin{figure}[!htb]
    \vspace{-0.2cm}
    \centering
    {\epsfig{file = 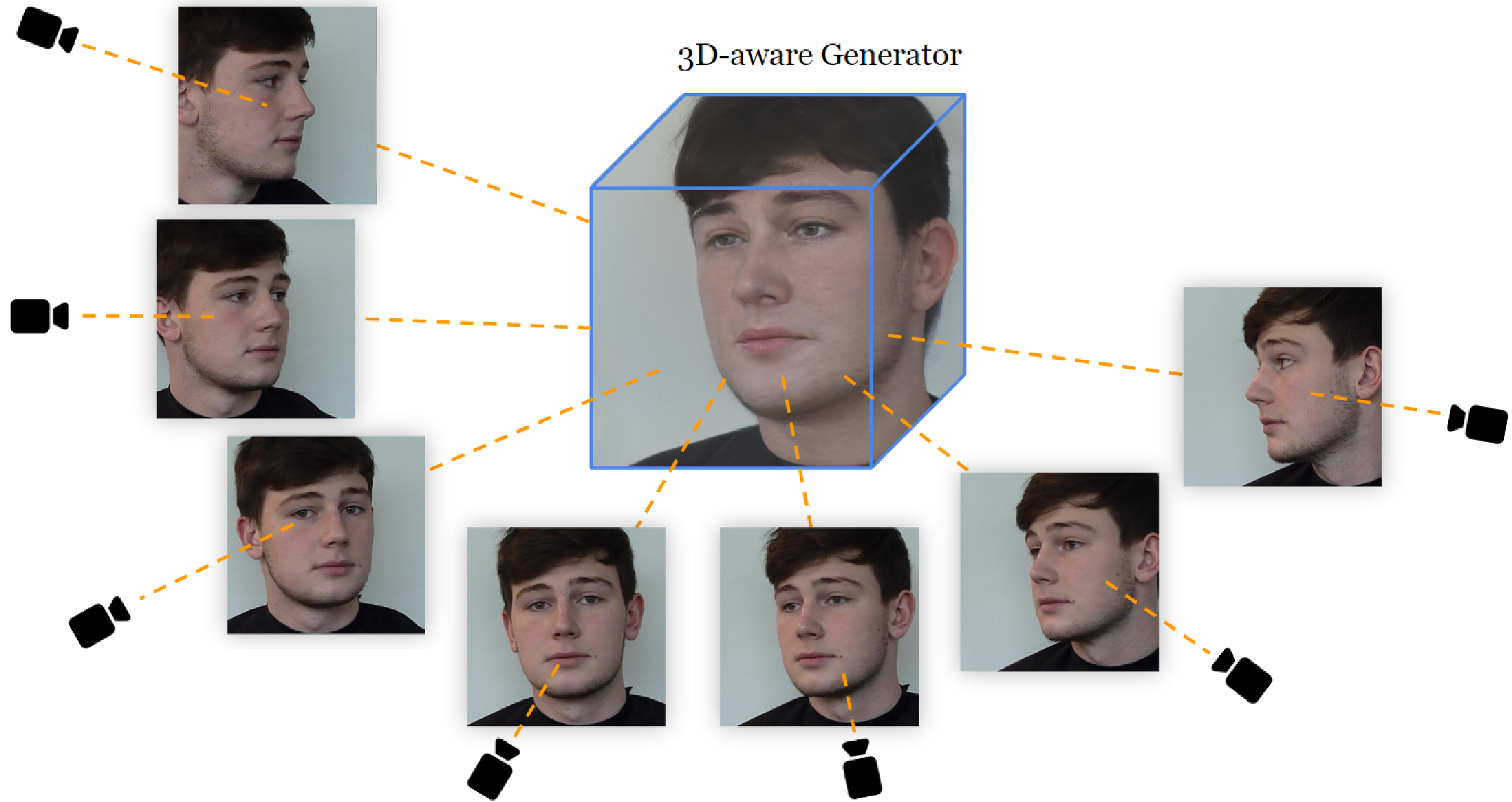, width = 7.5cm}}
    \caption{Example Multi-view optimization using 7 targets.}
    \label{fig:mult_opt}
\end{figure}

Using the target frames from all $N$ different views, we perform a $\mathcal{W+}$ optimization \cite{abdal_image2stylegan_2019} for all targets simultaneously. This is achieved by accumulating the gradients for all targets with respect to multiple loss terms and optimizing a single latent vector $w \in \mathcal{W+}$ using gradient descent. The loss functions we use are as follows: a perceptual loss $\mathcal{L}_P$ \cite{lpips}, a $\mathcal{L}_2$ pixel loss, an $ID$ loss \cite{arcface} and a regularization term $\mathcal{L}_r$ \cite{an_panohead_2023} that prevents the latent vector $w$ to diverge from the average latent vector in $\mathcal{W+}$. Altogether, this results in the following combined loss term:

\begin{equation}\label{eq3}
    \mathcal{L} = \sum_i^N \lambda_P \mathcal{L}_P^i + \lambda_2 \mathcal{L}_2^i + \lambda_r \mathcal{L}_r^i + \lambda_{ID} \mathcal{L}_{ID}^i.
\end{equation}

After optimizing $N$ separate latent vectors in $\mathcal{W+}$, we perform PTI using multiple targets. Analogously to $\mathcal{W+}$ optimization, we accumulate the gradients for all $N$ views and then fine-tune the weights of the generator. We apply the same loss function, however, without the $\mathcal{L}_r$ regularization.

Altogether, our multi-view optimization can be summarized with the following three steps:
\begin{enumerate}
    \item sample $N$ target images with evenly spaced camera angles from a face video,
    \item perform $\mathcal{W+}$ optimization with all $N$ target images simultaneously using gradient accumulation,
    \item perform PTI with all $N$ target images.
\end{enumerate}

\subsection{Multi-latent Optimization}
\label{multi_latent_opt}
When capturing a video sequence for 3D face reconstruction, it is advisable to minimize facial movements. Nevertheless, slight movements are unavoidable, e.g.~at the eyes or the jaw. 
This causes conflicts when seeking a single consistent 3D representation that matches all different poses. To address this problem, we introduce a multi-latent optimization process that simultaneously optimizes different latent vectors for different views. To achieve this, we first optimize a single latent vector as described in section \ref{multi-view-single-w}. Subsequently, we use the resulting vector to initialize $M$ new latent vectors $w_i \in \mathcal{W}+$. These latent vectors are then optimized separately, one for each viewpoint, using the same loss formulation as in section \ref{multi-view-single-w}. Also, we also use PTI afterwards. I.e.~we optimize the generator weights.

During inference, we interpolate between all fine-tuned latent vectors based on the camera angle. Specifically, we select the two latent vectors, $w_1$ and $w_2$, with the closest distance to the camera and apply linear interpolation in-between.

\begin{equation}\label{eq4}
    I = G(\textit{lerp}(w_1, w_2, c_{\textit{angle}}), c)
\end{equation}

Here, $c$ denotes the camera parameters during inference, whereas $c_{\textit{angle}}$ describes the normalized angle between $c_1$ and $c_2$, which are the camera parameters of $w_1$ and $w_2$ respectively.

\begin{figure}[!ht]
  \centering
    {\epsfig{file = 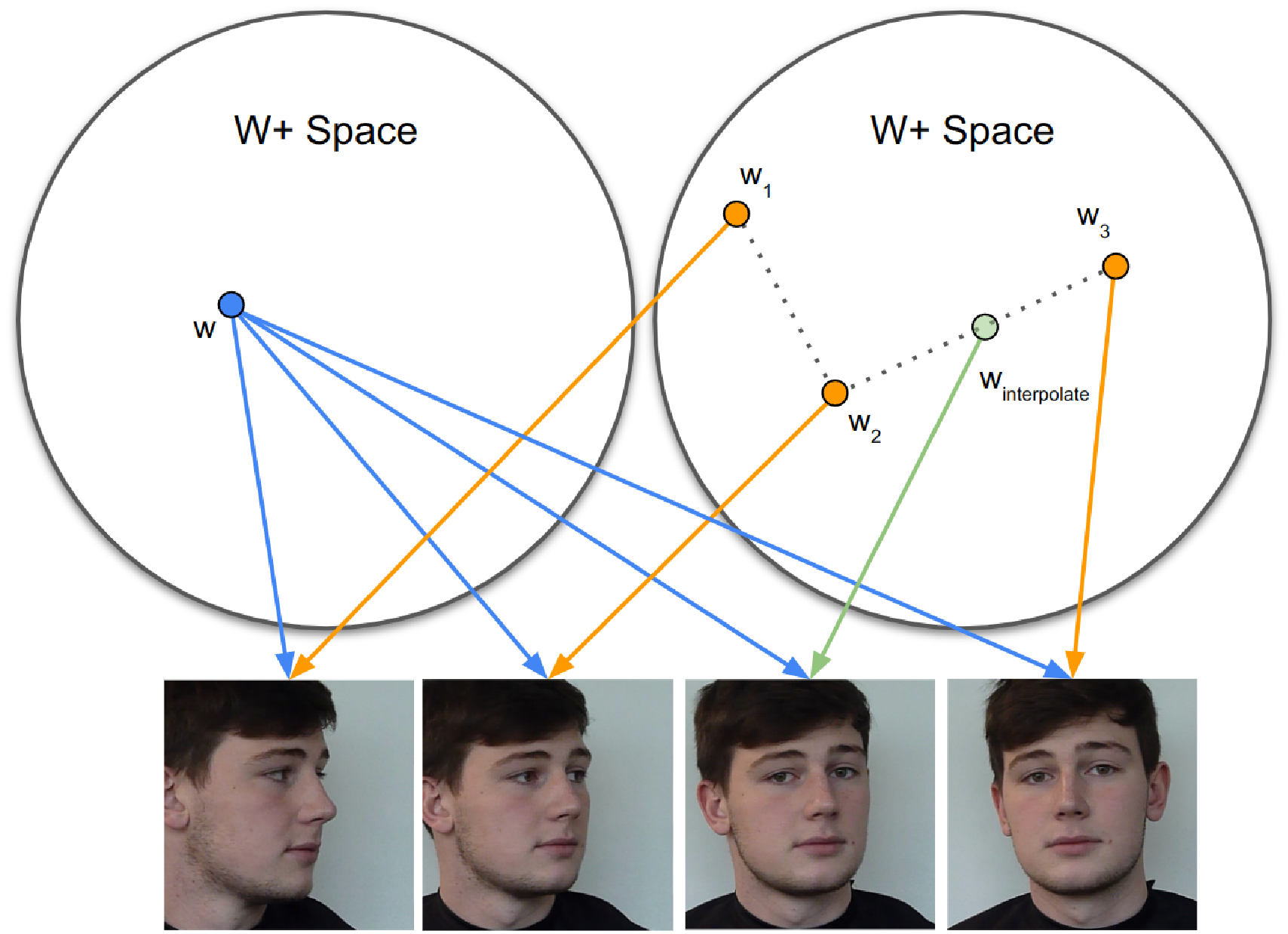, width = 7.0cm}}
  \caption{An illustration, highlighting the key difference between single-latent optimization (left) and multi-latent optimization (right). To transition between latent vectors, we apply linear interpolation in $\mathcal{W+}$.}
  \label{fig:wspace}
\end{figure}

With this approach, we facilitate the inversion process by eliminating the need to identify a single latent vector that represents all views simultaneously. This can be a challenging task, particularly for inconsistent input images from dynamic face videos. On the other hand, however, as all $M$ latent vectors are optimized independently, we have to deal with sudden changes when interpolating in-between neighboring latent vectors. To ensure seamless transitions, we apply three regularization methods, i.e.~a latent distance regularization, an interpolation regularization and a depth regularization, which are explained below.
\\

\noindent {\bf{Latent Distance Regularization:}} To preserve coherence, we incorporate a regularization term that penalizes the $L_2$ distance between all $M$ optimized vectors and the initial vector.

\begin{equation}\label{eq5}
    \mathcal{L}_{\textit{dist}} = \sum_{i=1}^M \lVert w_\text{i} - w_\text{init} \rVert^2
\end{equation}

This way, all latent codes are enforced to remain in close proximity to one another, thereby reducing the probability for traversing unknown regions when interpolating during inference.
\\

\noindent {\bf{Interpolation Regularization:}} To improve the image quality along the interpolation paths, we also optimize interpolated views during inversion. For this purpose, we sample a new latent vector centered between two adjacent latent vectors. Given this interpolated vector, we synthesize an image and apply a gradient step with respect to the loss functions mentioned in section \ref{multi-view-single-w}. As we create the intermediate vectors using linear interpolation, we  simultaneously optimize the position of both neighboring latent vectors. This implicitly creates a relation between all viewing points, similar to a doubly linked list. Visually, this is illustrated in figure \ref{fig:wspace}, where the green dot represents the interpolated vector centered in-between $w_2$ and $w_3$. Using this vector for optimization, implicitly optimizes $w_2$ and $w_3$.
As all neighboring vectors have to cooperate so that their interpolated vector yields good inversion quality, we expect that this approach enforces smoother interpolation paths.
\\

\noindent {\bf{Depth Regularization:}}
Finally, we add a regularization term to enforce the consistency of the geometry. As we allow each latent vector at each viewpoint to be unique, the geometry might show sudden changes when rotating during inference. In order to avoid such effects, we add a regularization that compares the depth maps for $M$ latent vectors when viewed from the same camera position. Specifically, we first calculate the average depth map for every camera viewpoint $c_i$ when synthesized with all $M$ latent vectors. Subsequently, the resulting average depth map for every camera viewpoint $c_i$ is then compared to the corresponding depth map created by the i-th latent vector $w_i$ using MSE. If all depth maps are the same, this loss returns 0. This can be formalized as follows:

\begin{equation}
\label{eq6}
    \mathcal{L}_{\textit{depth}}^i = \lVert G_{\textit{depth}}(w_i,c_i) - \frac{1}{M} \sum_{j=1}^M G_{\textit{depth}}(w_j,c_i) \rVert^2
\end{equation}

Here, $G_{\textit{depth}}$ denotes the differentiable depth image, synthesized by the generator.
In summary, multi-latent optimization follows three steps:

\begin{enumerate}
    \item perform multi-view optimization to initialize $M$ latent vectors,
    \item optimize each latent vector separately for a specific viewpoint using the consistency regularization methods,
    \item fine-tune the weights of the generator using PTI, with respect to all $M$ latent vectors.
\end{enumerate}

\begin{table*}[thp!]
    \caption{Quantitative comparison between single-view, multi-view and multi-latent inversion. All metrics on the left only consider the frontal image and all rotation metrics on the right are averaged over 180 images, ranging from the lowest viewing angle to the highest. We average our results over all six face videos from the dataset \cite{Colloff2022}. We compare our results to five single-view inversion methods: $\mathcal{W}+$ inversion \cite{abdal_image2stylegan_2020}, PTI \cite{roich_pivotal_2021}, TriplaneNet \cite{bhattarai_triplanenet_2023}, HFGI3D \cite{xie_high-fidelity_2023}, and SPI \cite{symetry_prior}.
    }
    \label{tab:compare} 
    \vspace{0.2cm}
    \centering
    \begin{adjustbox}{width=15.8cm,center}
    \begin{tabular}{|c|cccc|cccc|}
        \hline
        & \multicolumn{4}{|c|}{Single-view Metrics} & \multicolumn{4}{|c|}{Multi-view Metrics} \\
        Method & $MSE \downarrow$ 
        & $LPIPS \downarrow$ 
        & $\textit{MS-SSIM} \uparrow$ 
        & $ID\uparrow$ 
        & $MSE_R \downarrow$ 
        & $LPIPS_R \downarrow$ 
        & $\textit{MS-SSIM}_R \uparrow$ 
        & $ID_R \uparrow$\\
        \hline
        \textbf{Single-view} & & & & & & & & \\
        $\mathcal{W+}$ Opt      & 0.008	&	0.16	&	0.92	&	0.68	& 0.071	&	0.37	&	0.74	&	0.45 \\
        PTI         & \textbf{0.002}	&	\textbf{0.08}	&	\textbf{0.97}	&	0.93	& 0.075	&	0.36	&	0.75	&	0.62 \\
        TriplaneNet & 0.005	&	0.22	&	0.96	&	0.90	& 0.128	&	0.45	&	0.52	&	0.69 \\
        HFGI3D      & 0.004	&	0.17	&	0.94	&	0.68	& 0.155	&	0.39	&	0.66	&	0.44 \\
        SPI         & 0.016	&	0.14	&	0.90	&	\textbf{0.95}	& 0.073	&	0.37	&	0.74	&	0.65 \\
        \hline
        \textbf{Multi-view (ours)} & & & & & & & & \\
        3 Views & 0.014	&	0.18	&	0.88	&	0.68	& 0.045	&	0.29	&	0.79	&	0.61 \\
        5 Views & 0.018	&	0.19	&	0.89	&	0.73	& 0.024	&	0.24	&	0.85	&	0.68 \\
        7 Views & 0.016	&	0.18	&	0.89	&	0.78	& 0.022	&	0.22	&	0.87	&	0.73 \\
        9 Views & 0.014	&	0.17	&	0.90	&	0.80	& 0.021	&	0.21	&	0.88	&	0.75 \\
        \hline
        \textbf{Multi-latent (ours)} & & & & & & & & \\
        9 Latents   & 0.007	&	0.15	&	0.94	&	0.89	& 0.010	&	0.19	&	0.92	&	0.83 \\
        + Dist. Reg & 0.008	&	0.15	&	0.94	&	0.87	& 0.010	&	0.19	&	0.92	&	0.82 \\
        + Inter. Reg& 0.007	&	0.14	&	0.95	&	0.88	& 0.009	&	0.18	&	0.93	&	0.85 \\
        + Depth Reg & 0.007	&	0.14	&	0.95	&	0.88	& \textbf{0.009}	&	\textbf{0.18}	&	\textbf{0.93}	&	\textbf{0.86} \\
        \hline
    \end{tabular}
    \end{adjustbox}
\end{table*}

\section{\uppercase{Experiments}}
\label{sec:experiments}

\subsection{Configuration}
The configuration for our experiments closely follows the 3D inversion implementation of \cite{an_panohead_2023}. In particular, we optimize the latent vector for 500 iterations using a decaying learning rate starting from 0.1. We have also performed longer experiments with more optimization steps; however, they do not show better results. To weight our four loss functions, we set $\lambda_p=1.0$, $\lambda_2=0.1$, $\lambda_r=1.0$ and $\lambda_{ID}=1.0$.
In order to initialize the latent vectors for our multi-latent experiments, we use a prior multi-view optimization with seven target views.

In all experiments, we use an EG3D \cite{chan_efficient_2022} generator that was pre-trained on the large-pose Flickr face (LPFF) dataset \cite{wu_lpff_2023}. Compared to FFHQ \cite{karras_progressive_2018}, LPFF adds additional images from large viewing angles, which facilitate 3D inversion from multiple views. For the comparisons with TriplaneNet, HFGI3D and SPI, however, we follow the official implementations, which use FFHQ.

\subsection{Data}
For our face videos, we use a dataset that imitates UK police lineup videos \cite{Colloff2022}. Those videos show a person turning their head from one side to the other. For six videos, we extract all frames and perform the same data pre-processing as in \cite{chan_efficient_2022}.

\subsection{Metrics}
In order to validate the performance of our inversion method, we apply the following four commonly used metrics: MSE, LPIPS \cite{lpips}, MS-SSIM \cite{ms_ssim} and ID similarity \cite{arcface}. Instead of only comparing the frontal views, we select 180 target frames with evenly distributed viewing angles across the maximal range and calculate the average score, respectively. We will refer to those rotation metrics as $\textit{MSE}_R$, $\textit{LPIPS}_R$, $\textit{MS-SSIM}_R$ and $\textit{ID}_R$.

\subsection{Quantitative Results}
We compare our results to three state-of-the-art 3D GAN inversion methods, i.e.~$\mathcal{W}+$ inversion \cite{abdal_image2stylegan_2019}, PTI \cite{roich_pivotal_2021}, TriplaneNet \cite{bhattarai_triplanenet_2023}, HFGI3D \cite{xie_high-fidelity_2023} and SPI \cite{symetry_prior} 

In the following, we will first examine the results of the single-view 3D inversion, which only uses a frontal face image to reconstruct a 3D model. After that we will demonstrate the results when applying our multi-view inversion method that optimizes a single latent vector for multiple target views simultaneously. And finally, we will demonstrate the results of our multi-latent inversion method, in which we optimize multiple latent vectors, one for each viewpoint, separately.
\\

\begin{figure}[!ht]
  \centering
  {\epsfig{file = 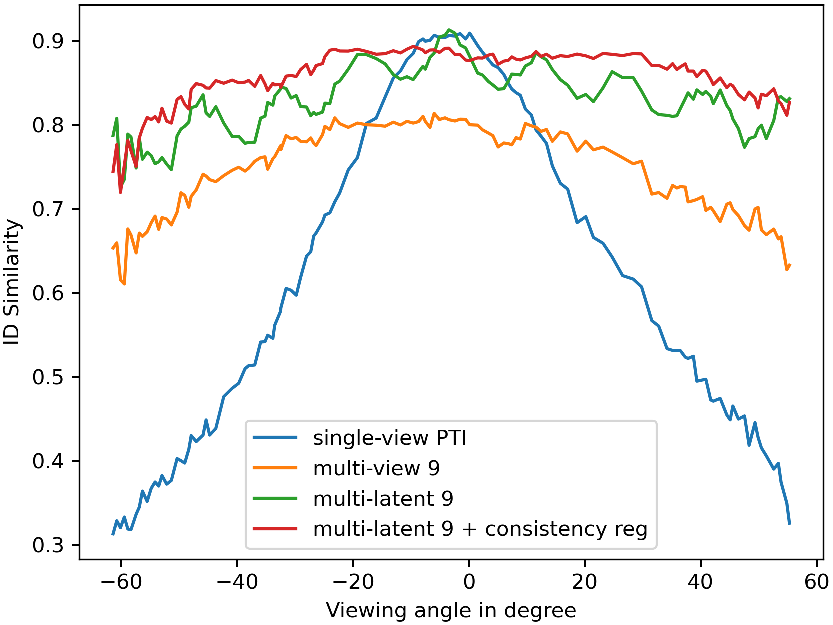, width = 7.5cm}}
  \caption{Average ID similarity across all camera viewpoints (higher is better). At 0°, the face is shown from a frontal view.}
  \label{fig:angle_graph}
\end{figure}

\noindent {\bf Single-view:} 
As expected, the single-view experiments perform very well for all methods, when it comes to synthesizing the face from a frontal viewpoint. Here, we find the overall best performance using PTI. 
However, for the evaluation of rotated views, the rotation metrics show considerably worse performance. This underlines that single-view 3D GAN inversion misses significant information to reconstruct the true geometry of a face. In figure \ref{fig:angle_graph}, we plot the ID similarity against the viewing angle of the camera. For the single-view PTI experiment (blue line), we observe the highest ID similarity at the frontal view, whereas the similarity decreases as soon as we rotate the camera in either direction.

\noindent {\bf Multi-view:} The results of the multi-view experiments, on the other hand, are contrary to the single-view results. The rotation metrics score significantly better, although at the cost of single-view quality. This can also be seen in the ID similarity graph in figure \ref{fig:angle_graph} (orange line). Here, we observe worse quality when generating a frontal view, compared to the single-view experiment. However, for angles that are more than 20 ° apart from the center, the ID similarity is improved. 

In our experiments, we test the multi-view inversion with different amounts of target images $N$. Generally, we find that both single-view metrics and rotation metrics favor experiments with more target images. This is shown in table \ref{tab:compare}.
\\

\noindent {\bf Multi-latent:} Using our multi-latent optimization, the algorithm does not have to identify a single latent vector that best represents all views simultaneously. As highlighted in table \ref{tab:compare}, this results in the overall best inversion quality for the rotation metrics. Additionally, the single-view metrics show results similar to those of the single-view optimization experiments. As demonstrated in figure 
\ref{fig:angle_graph} (red line), we observe the highest ID similarity throughout a full rotation. 

In figure \ref{fig:angle_graph} (green graph), we also compare the multi-latent experiment without any consistency regularization. Here, we observe a spike at every location for which we provided the optimization algorithm with a target image. This is no longer the case after applying all our three consistency regularization methods. Instead, we observe a smooth and consistent quality distribution. This underlines that the regularization methods help to improve the inversion quality along the interpolated paths in-between neighboring latent vectors.

\begin{figure*}[!ht]
  \centering
   {\epsfig{file = 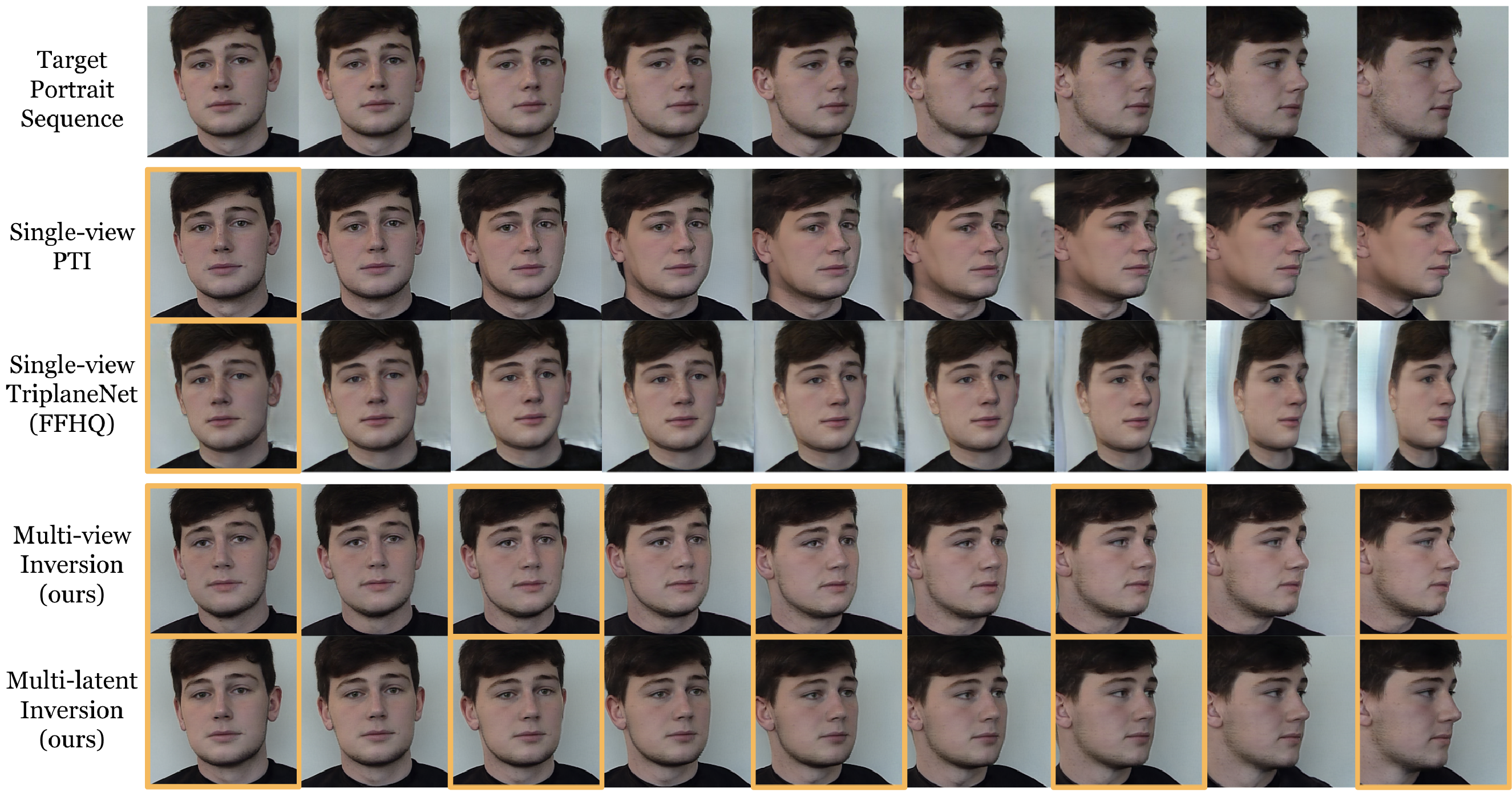, width = 15.8cm}}
  \caption{Qualitative comparison between state-of-the-art single-view inversion methods \cite{roich_pivotal_2021} (row 2),  
  \cite{bhattarai_triplanenet_2023}
  (row 3) and our multi-view and multi-latent inversion methods (rows 4 \& 5). The orange boxes indicate, whether the corresponding target image from row 1 was used during optimization. }
  \label{fig:compare_11}
\end{figure*}

All ID similarity curves in figure \ref{fig:angle_graph} show slightly worse inversion quality when the viewing angle moves from the center. We believe that this is due to the training data distribution, which includes more faces from a frontal viewpoint.
\\

\noindent {\bf 3D Consistency:} As our method allows different latent vectors for different viewpoints, we receive slightly different geometries during a full rotation. To reduce this effect, we implemented three regularization methods: latent distance regularization, interpolation regularization, and depth regularization. To test their effectiveness, we propose a consistency metric that measures the average standard deviation for all depth maps when synthesized from different latent vectors at the same camera viewpoint. For this purpose, we select $M=60$ viewpoints, along with $M$ interpolated latent vectors (using equation (\ref{eq4})) and compute the standard deviation of all $M$ depth maps for each viewpoint. Given all $M$ standard deviations, we then compute the average:

\begin{equation}
\label{eq7}
    \text{Depth Consistency} = \frac{1}{M} \sum_{i=1}^M \sigma_i
\end{equation}

\begin{equation}
\label{eq8}
    \sigma_i = \sqrt{\frac{1}{M} \sum_{j=1}^M  (G_{\textit{depth}}(w_j,c_i) - \mu_i)^2}
\end{equation}

\begin{equation}
\label{eq9}
    \mu_i = \frac{1}{M} \sum_{j=1}^M  G_{\textit{depth}}(w_j,c_i)
\end{equation}

\begin{table}[htbp]
    \caption{Results of the depth consistency and latent distance for all multi-latent experiments.}
    \label{tab:depth_consistency} 
    \centering
    \begin{tabular}{|c|c|c|c|}
        \hline
        Experiment & Depth Cons. $\downarrow$ & Latent Dist. $\downarrow$ \\
        \hline
        Multi-latents 9 & 0.023         & 76.35 \\
        + Dist. Reg     & 0.018         & 16.14 \\
        + Inter. Reg    & 0.020         & \textbf{11.06} \\
        + Depth Reg     & \textbf{0.014} & 11.18 \\
        \hline
    \end{tabular}
\end{table}

The result of the depth consistency metric is listed in table \ref{tab:depth_consistency}. It shows that the latent distance regularization and the depth regularization both improve the geometric consistency of the 3D inversion. The interpolation regularization, on the other hand, slightly decreases the consistency. This is expected as the interpolation regularization mainly focuses on inversion quality along the interpolated path instead of establishing a geometric consistency.

In addition to depth consistency, we also measure the average distance between neighboring latent vectors in $\mathcal{W}+$. Here, we observe that the distance is considerably reduced when applying both distance regularization and interpolation regularization. This underlines that both regularization methods help the inversion algorithm to find a more compact solution, where all latent vectors are close to each other.

\begin{figure*}[!ht]
  \centering
    {\epsfig{file = 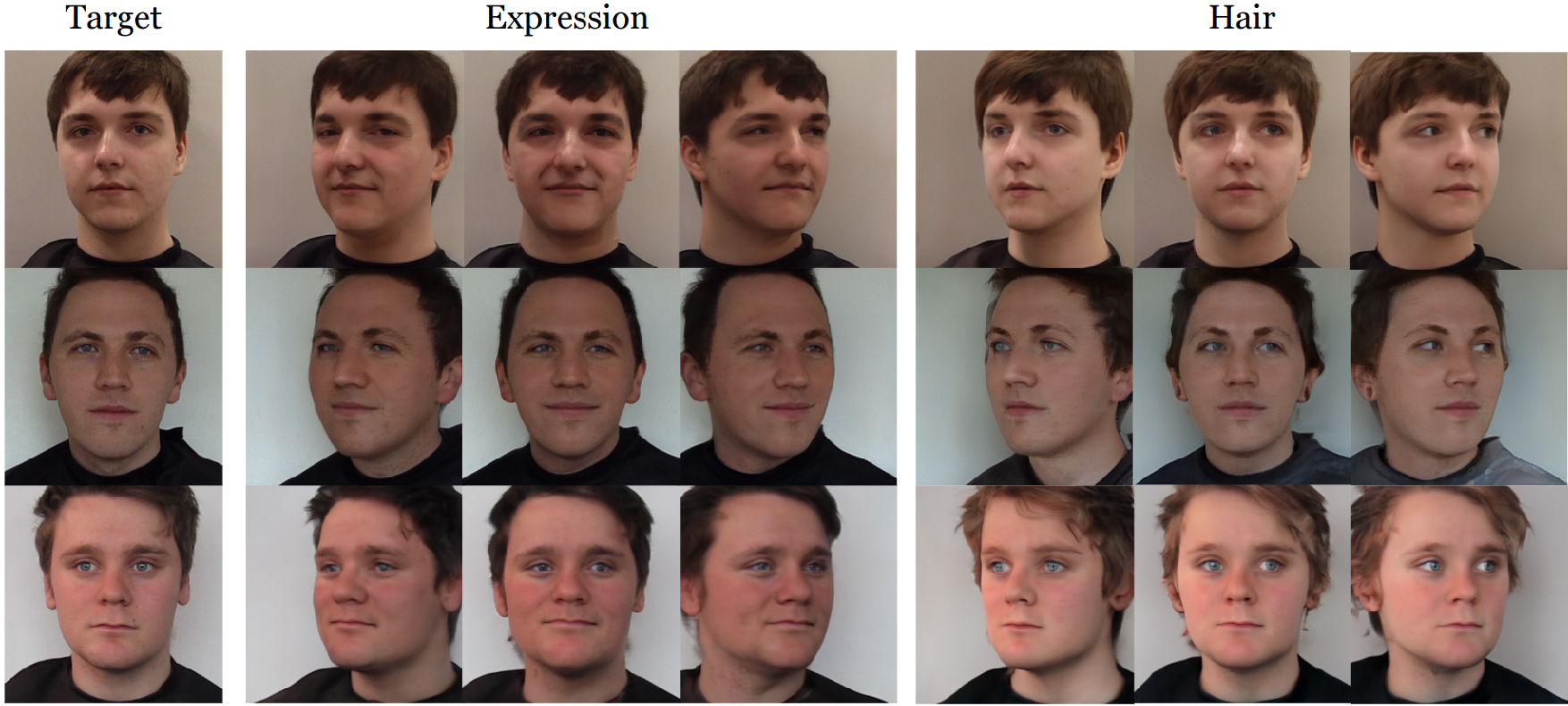, width = 15.8cm}}
   
  \caption{We first invert a face using multi-latent inversion and then edit the result using GAN Space \cite{harkonen_ganspace_2020}. We move all latent vectors along the seventh and and fourth PCA componant to edit the expression and hair, respectively.}
  \label{fig:edit}
\end{figure*}

\subsection{Qualitative Results}
Figure \ref{fig:compare_11} underlines the inversion improvements qualitatively. We compare both single-view experiments (PTI and TriplaneNet) to our multi-view and multi-latent approaches. As the single-view experiments have to invent 
information for views showing the side, we observe a considerably worse quality when rotating the camera from the center. 

The multi-view experiment, on the other hand, shows a very consistent inversion quality throughout. Nevertheless, for steep viewing angles, we observe clear discrepancies in geometry and shading.

The best inversion quality is obtained using our multi-latent approach. It shows very similar renderings to the target images and also provides good results for viewpoints that were not present during optimization.

\subsection{Editing Results}
Finally, we demonstrate the editing capabilities of the inverted images. We use our multi-latent inversion algorithm and edit the resulting image using GAN Space \cite{harkonen_ganspace_2020}. This method first calculates the PCA components of $\mathcal{W}+$. Afterwards, we move all $M$ latent vectors in the direction of specific components. Since the components correlate with certain image features, we are able to control the appearance of the subject. In figure \ref{fig:edit}, we demonstrate this with three examples where we change the expression and the hair. This can be achieved by amplifying the seventh or third PCA component of the latent vectors.

Figure \ref{fig:edit} underlines that the edit is invariant to the viewing direction. Subsequently, all $M$ latent vectors behave identically when moving in the latent space. Nevertheless, the GAN Space method does not allow very precise and controlled editing of the images. For instance, when changing the hair in figure \ref{fig:edit}, we observe other attributes such as the skin or the eyes, change as well. To address this limitation, other editing methods, such as DragYourGAN \cite{pan_drag_2023} or StyleFlow \cite{styleflow} could be applied instead.

\section{\uppercase{Conclusion}}
\label{sec:conclusion}

In our work, we introduce a novel 3D GAN inversion method that incorporates the information from all viewing angles of dynamic face videos to accurately synthesize 3D renderings from any perspective. We show that our method significantly outperforms current state-of-the-art inversion methods, which all rely on re-synthesizing a full 3D model using only one image showing a face from a frontal viewpoint. In addition, we introduce several regularization methods to improve the consistency of the 3D renderings when viewed from various viewpoints. Finally, we demonstrate the editing capabilities when utilizing our 3D inversion method.

For future work, we will investigate how well hybrid- or encoder-based inversion methods could be combined with multi-view or multi-latent inversion. This could potentially speed up the optimization process considerably. In addition, we will test how well other GAN editing methods could be applied after re-synthesizing an image via multi-latent inversion.
\newpage

\section*{\uppercase{Acknowledgements}}
This work has partly been funded by the German Research Foundation (project 3DIL, grant no.~502864329) and the German Federal Ministry of Education and Research (project VoluProf, grant no.~16SV8705).

\bibliographystyle{apalike}
{\small
\bibliography{example}}


\end{document}